\documentclass{bmvc2k}

\title{\hspace{1.2em} Zero Shot Domain Generalization}

\addauthor{Udit Maniyar}{es16btech11024@iith.ac.in}{1}
\addauthor{Joseph K J}{cs17m18p100001@iith.ac.in}{1}
\addauthor{Aniket Anand Deshmukh}{Aniket.Deshmukh@microsoft.com}{2}
\addauthor{Urun Dogan}{urundogan@gmail.com}{2}
\addauthor{Vineeth Balasubramanian}{vineethnb@iith.ac.in}{1}

\addinstitution{
 Indian Institute of Technology, \newline Hyderabad,
 India
}
\addinstitution{
 Microsoft, Sunnyvale, CA, USA
}

\runninghead{Udit, Joseph, Aniket, Urun, Vineeth}{Zero Shot Domain Generalization}

\def\etal{\emph{et al}\bmvaOneDot}
\usepackage{times}
\usepackage{epsfig}
\usepackage{graphicx}
\usepackage{amsmath}
\usepackage{amssymb}

\usepackage{booktabs}
\usepackage{algorithm}
\usepackage{amsmath,amssymb,amsfonts}
\usepackage{mathtools}
\usepackage[super]{nth}
\usepackage{wrapfig, blindtext}
\usepackage{float}

\usepackage[noend]{algpseudocode}
\usepackage{lscape}

\newcommand{\ourmethod}{Semantic FC (S-FC)}
\newcommand{\mtaezsdg}{Semantic MTAE (S-MTAE)}
\newcommand{\aggzsdg}{Semantic AGG (S-AGG)}

\begin{document}

\maketitle

\begin{abstract}

Standard supervised learning setting assumes that training data and test data come from the same distribution (domain). Domain generalization (DG) methods try to learn a model that when trained on data from multiple domains, would generalize to a new unseen domain. We extend DG to an even more challenging setting, where the label space of the unseen domain could also change. We introduce this problem as \textit{Zero-Shot Domain Generalization} (to the best of our knowledge, the first such effort), where the model generalizes across new domains and also across new classes in those domains.
We propose a simple strategy which effectively exploits semantic information of classes, to adapt existing DG methods to meet the demands of Zero-Shot Domain Generalization.
We evaluate the proposed methods on CIFAR-10 \cite{krizhevsky2009learning}, CIFAR-100 \cite{krizhevsky2009learning}, F-MNIST \cite{xiao2017fashion} and PACS \cite{li2017deeper} datasets, establishing a strong baseline to foster interest in this new research direction.
\end{abstract}

% In this paper, we formalize the problem setting and propose baseline methods based on existing domain generalization literature. We then introduce a novel methodology to solve Zero Shot Domain Generalization. We show how semantic embedding helps in better generalization. 
% and show promising results

\section{Introduction}

\begin{figure}[t]
  \centering
% \resizebox{1\textwidth}{!}{%

  \includegraphics[width=0.75\textwidth]{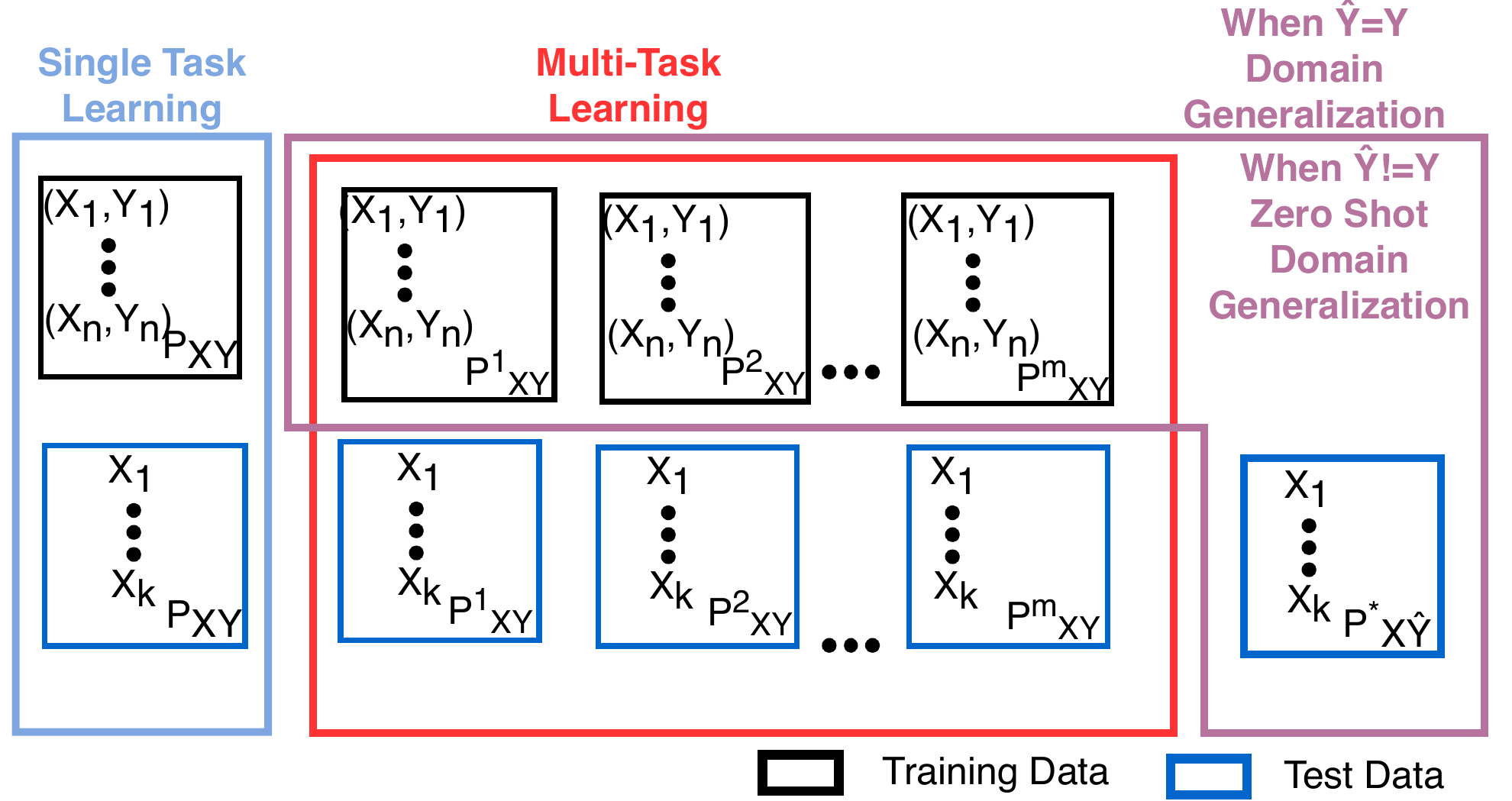}
% }
\caption{\footnotesize{
In a single-task learning (light blue), both training data-points and test data-points comes from the same distribution $P_{XY}$. In multi-task learning (red), data from all distributions are available for training ($P^{1}_{XY}\cdots P^{m}_{XY}$). In domain generalization (pink), the model is evaluated on data from unseen distribution $P^{*}_{XY}$, while in Zero-Shot DG, the model needs to be performant on any unseen $P^{*}_{X\hat{Y}}$, where $\hat{Y} \neq Y$.}}
\label{fig_intro}
\vspace{-10pt}
\end{figure}

Generalization is a key desideratum for machine learning models to scale to the dynamic nature of the real world.  
The standard supervised learning framework assumes that train and test data are from the same distribution (domain).
Domain generalization techniques \cite{blanchard:11:nips,ghifary2015domain,li2017deeper,li2017learning,li2019feature} demand to train a model in such a way that it can generalize to a novel domain at inference, by gracefully handling domain shift. However, current domain generalization methods assume the same classes to be present in all domains (including unseen test domains), which is a restriction on the application of such methods. Our work attempts to relax this assumption, and allow novel test domains to have new classes that were not present in any training domain. We introduce this harder problem as \textit{Zero-Shot Domain Generalization}, and to the best of our knowledge, is the first such effort (Fig \ref{fig_intro} illustrates the setting). We note that the standard zero-shot learning problem \cite{labelembedding2015,deepzerotexticcv,DEVISE_NIPS2013_5204,socher2013zero,goodbadugly_2017} provides a model to generalize to unseen classes, but assumes that datapoints come from a single known domain.

We hypothesize that learning a domain-invariant feature representation, with explicit class information, would help address Zero-Shot Domain Generalization. Encoding class information in the feature space ensures smooth transition of information from the classes that are seen during training on multiple source domains, to the unseen set of classes in the new domain. To this end, we adapt state-of-the-art domain generalization methods - in particular, Feature Critic Network (FC)~\cite{li2019feature} and Multi-Task Auto Encoder (MTAE)~\cite{ghifary2015domain} - to ensure that their intermediate feature representations are semantically consistent. Our experimental evaluation on domain generalization variants of CIFAR-10 \cite{krizhevsky2009learning}, CIFAR-100 \cite{krizhevsky2009learning}, F-MNIST \cite{xiao2017fashion} and PACS \cite{li2017deeper} provide promise, and establish baselines for this new research problem.

\vspace{-10pt}
\section{Related Work}\label{sec:related_work}
\vspace{-3pt}
\paragraph{Domain Generalization:}

Multi-Domain Learning, Domain Generalization, Domain Adaptation are closely related topics. All of these deal with scenarios in which a model is trained on a source distribution and is used in the context of a different (but related) target distribution. In Multi-Domain Learning the target domain and train domains are the same. In Domain Generalization the target domain is different from that of train domains. In Domain Adaptation also the target domain is different from the train domain but we have access to unlabeled data from the target domains during training.

In domain generalization (DG), we are given training data from different domains and the objective is to generalize to a novel domain. Blanchard \etal 2011 \cite{blanchard2017domain,blanchard:11:nips} proposed a kernel mean embedding based algorithm to get similarity between different domains and transfer the learning. The same kernel mean embedding idea was used to extend this DG framework to a multiclass setting in \cite{deshmukh2017multiclass}. Ghifary \etal 2015 \cite{ghifary2015domain} proposed an autoencoder based method called Multi-Task Auto Encoder (MTAE) to learn domain-invariant features.
In MTAE, encoder for all domains is same but decoder is different which forces encoder to learn a common feature space for all domains. Ghifary \etal 2017 \cite{ghifary:2017:pami} proposes scatter component analysis (SCA) to solve domain generalization problem. SCA finds representation that trades between maximizing the separability of classes and data, and minimizing the mismatch between domains through scatter.
Motiian \etal 2017 \cite{motiian2017unified} proposed classification and contrastive semantic alignment (CCSA) to learn a domain-invariant embedding by minimizing the sum of classification loss, confusion alignment loss, and semantic alignment loss.
Confusion alignment loss is a distance between distribution and semantic alignment loss makes sure that samples from different domains and with different labels are mapped as far apart as possible in the embedding space.
Li \etal 2017 \cite{li2017deeper} proposed a method that takes advantage of the robustness of deep learning models to domain shift, and developed a low-rank parameterized CNN model for end-to-end DG learning. Li \etal 2017 \cite{li2017learning} proposed a meta-learning approach to solve for DG by simulating train-test domain shifts during training by synthesizing virtual test domains. Li \etal 2019 \cite{li2019feature} extended this approach to produce a more general feature extractor that can be used with any classifier, by simultaneously learning an auxiliary loss function that trains the feature extractor for improved domain invariance. None of these efforts attempted Zero-Shot DG, setting our work apart.\\
% Zero Shot learning is a supervised learning method, in which the classes seen in training data is disjoint from the classes present in the test data. In Zero shot learning we assume that only unseen classes are present in test data-set. This assumption has hindered the practical application of these methods and hence generalized zero-shot learning was introduced where the search space of classifier is both seen and unseen classes.

\noindent \textbf{Zero-Shot Learning (ZSL):}
On the other hand, Zero-Shot learning is a supervised learning method, in which the classes of images seen during training is disjoint from the classes present during testing. Successful zero-shot models should maintain very high accuracy when presented test examples from those unseen classes. 
In generalized zero-shot setting, the model should maintain good performance on all the classes which includes both the base classes and new zero-shot classes. The main idea in most zero-shot algorithms is to learn a semantically consistent correlation between seen and unseen classes, using semantic information between classes.
% This helps to transfer  knowledge from training data (seen classes) to test data (unseen classes). 
Semantic information can be either manually specified as attributes \cite{attributes2012cvpr,onlineincrementalzeroshot,NIPS2007_3217} or imparted from  word embeddings such as Word2Vec \cite{wor2vec_nips2013} or GloVe vectors\cite{pennington2014glove}.
Since attribute based methods use binary mapping they result in loss of information and hence have been shown to be suboptimal\cite{labelembedding2015}.
Semantic embedding based algorithms first learn an embedding function which maps visual space to semantic space, to aid in this task.
Socher \etal 2013 \cite{socher2013zero} minimizes Euclidean distance loss with word vectors in semantic space as objective, and learns a two-layer neural network for classification. Attribute Label Embedding (ALE) \cite{labelembedding2015}, Deep Visual Semantic Embedding (DEVISE) \cite{DEVISE_NIPS2013_5204} and Structured Joint Embedding (SJE) \cite{sje:structuraljointembedding}
learn a bilinear compatibility function to map image to semantic space. To learn the compatibility function, different objectives functions have been explored: DEVISE\cite{DEVISE_NIPS2013_5204} uses pairwise ranking objective, \cite{deepzerotexticcv} takes a dot product between the embedded visual feature and semantic vectors considering three training losses, including a binary cross entropy loss, hinge loss and Euclidean distance loss. Yongqin \etal 2017 \cite{goodbadugly_2017} presents a survey of zero-shot learning methods. 

% Most of these methods have good performance in zero-shot setting while they fail in generalized zero-shot setting. Novelty detection methods and outlier detection methods \cite{socher2013zero} are used to alleviate this to generalized zero-shot setting. 

Even though both DG and ZSL have been explored by the community in isolation, to the best of our knowledge, no existing work in literature has attempted to solve the Zero-Shot DG problem. This setting can have practical use in applications such as robotics, medical image analysis and general scene understanding. We identify this problem in our work, formalize the same and provide a methodology that can serve as a baseline.

%challenge here is that the model should be able to classify object to unseen classes, in an unseen domain. 
%Despite its wide practically appealing applications in lot of settings including robotics, medical image analysis and general scene understanding, the problem has been left unexplored. 
% We show that semantic embeddings helps in solving the domain generalization problem in a better way and hence also solving the Zero Shot Domain Generalization.

\vspace{-13pt}
\section{Zero-Shot Domain Generalization}
% We introduce Zero Shot domain generalization which is one level harder problem compared to Domain Generalization and Zero Shot learning.
% Our goal is to learn a model which has generalization performance both horizontally (across novel domains) and vertically (across novel classes).
% Since the problem is novel we define the problem, introduce new methods to solve the problem based on previous work done in this field, test the algorithms extensively on 4 datasets.
% \subsection{Problem Definition} \label{sec:formal_problem_statement}
\vspace{-4pt}
\subsection{Problem Definition}
% \noindent \textbf{:}
We define a domain as the joint distribution of feature and label space. Let the training data for the $i^{th}$ domain be:  \((X_{ij},Y_{ij}) \sim P_{XY}^i\) and \(P_{XY}^i \sim \mu\).  In a DG setting, the test data comes from the same distribution $\mu$:  $(X_j^T, Y_j^T) \sim P^T$ and \(P^T \sim \mu\), while in zero-shot domain generalization, it comes from a different distribution $\nu$:  $(X_j^T, Y_j^T) \sim P^T$ and \(P^T \sim \nu\), where  $\nu$ contains additional unseen labels compared to $\mu$ (note that feature space and common label space follow the same data generation structure as $\mu$ but it differs only for and when unseen classes are seen). 
%(corresponding marginal distribution of feature space and common label space between train and test domains is also same).

% We define a domain as the joint distribution of feature and label space. Let the training data for the $i^{th}$ domain be:  \((X_{ij},Y_{ij}) \sim P_{XY}^i\) and \(P_{XY}^i \sim \mu\).During testing, in DG and Zero-Shot DG, features and seen labels are drawn from the same distribution \(\mu\) as in training.

We assume all $(X,Y)$ pairs are drawn i.i.d. from their respective distributions.
% , and that $P_1, \ldots, P_N, P^T$ are iid from $\mu$.
% Let $N$ be the number of domains in training data-set and
In particular, let $Y^{tr}$ and $Y^{ts}$ represent the set of classes in training and test data respectively, such that $Y^{tr} \cap Y^{ts} = \emptyset$.
The training data for the $i^{th}$ domain is given by $ D_i = \{X_{ij}, Y_{ij}\}_{1 \leq j \leq n_i}$ and $Y_{ij} \in Y^{tr}$, and the test data set be $D^T = \{X_{j}^T, Y_j^T\}_{1 \leq j \leq n_T}$ and $Y^{T}_{j} \in Y^{ts}$ where $n_{i}$ is the number of images in the $i^{th}$ domain.
The main objective of ZSDG problem is to train a model on all training domains $D = \{D_1, D_2, \ldots D_N \}$ to perform well on $D^T$.
% The main objective of Zero Shot Domain Generalization is to train a model on all the training domains $D = \{D_1, D_2, \ldots D_N \}$ such that the resultant model performs well on $D^T$, which would contain novel classes.

% \subsection{Importance of Semantic Embeddings}

% In zero-shot learning semantic embeddings are popularly used  generalize over novel classes. We claim that using semantic embedding also helps in solving the domain generalization problem by providing a new modality to the data. By using the semantic embedding and restricting the lower level features of the model to the semantic space a shared invariant representation is learned where semantic alignment is accounted for.We make an inherent assumption that classes which appear visually similar should also be semantically similar. Thus using semantic space helps us in visual classification task.

% \begin{itemize}
%     \item How semantic restriction helps in Domain Generalization
%     \item How semantic restriction helps in Zero Shot learning.
    
% \end{itemize}
\vspace{-10pt}
\subsection{Proposed Approach}
% \noindent \textbf{Proposed Approach:}
We propose a generic approach to extend the state-of-the-art DG techniques to solve zero-shot domain generalization. The key insight is to bind the intermediate domain-invariant feature representations to a semantic space that is shared across the seen classes of the old domains and the unseen classes of the new domain.
Such use of semantic space has been successfully used in recent ZSL methodologies \cite{goodbadugly_2017,socher2013zero,DEVISE_NIPS2013_5204}. 

Existing DG methods~\cite{ghifary2015domain,li2019feature} can be considered as a composition of a feature extractor function $f_{\theta}$ and classifier function $g_\phi$: $(g\circ f)( \mathcal{\boldsymbol I})$, where $\mathcal{\boldsymbol I}$ is the input data-point. In this paper, we only consider DG methods which generate domain invariant features and train a common classifier on those features. We restrict domain invariant features to semantic space, forcing the model to be semantically consistent along with domain invariance. Our proposed method can be extended to any DG method which learn domain invariant features but does not apply to methods which learn different features for different domains. 
While training the proposed semantically constrained DG approach, the features
% $\boldsymbol{F}$
generated by $f_{\theta}$ are projected to a semantic space of labels. Images of similar classes are grouped together, and dissimilar classes spaced apart by a semantic alignment loss given in Equation \ref{eqn:semantic_loss}. 
% \begin{figure}[h]
%   \centering
%   \includegraphics[width=0.48\textwidth]{latex/Flowchart.pdf}
% \caption{Flowchart}
% \label{fig:arch}
% \end{figure}
By using the semantic embedding and restricting lower-level features of the model to a semantic space, a shared invariant representation is learned where semantic alignment is accounted for.
We make an inherent assumption that classes which appear visually similar should also be semantically similar (as in other ZSL methods). Thus using semantic space helps us in the visual classification task. We use word embeddings of classes - in particular, simple GloVe embeddings \cite{pennington2014glove} trained on Common Crawl corpus - as the semantic space in this work. One could use more complex embedding functions to study this even further.

During inference, we depart from traditional DG methods which train a separate classifier (neural network or SVM) on the domain-invariant features. We instead use nearest neighbour (NN) search in the semantic space. Our method is lightweight and can scale to large numbers of unseen classes using existing efficient NN search methods. We now describe how we infuse semantic information into three DG methods, to solve zero-shot DG.

Our approach focuses on DG methods which learn domain-invariant features and cannot be any combination of DG + ZSL methods. For example existing SOTA ZSL methods, based on generative models, cannot be directly extended this way. The proposed method is novel in terms of how one does training + inference in ZSDG, diff from existing DG methods.

\vspace{-10pt}
\subsubsection{\aggzsdg}
% \textbf{\aggzsdg:} 
Aggregation (AGG) is a simple baseline, which has strong DG performance \cite{li2017deeper, li2019feature}. Here, we group data from all domains and train the network on this multi-domain dataset. In line with our generic definition for DG functions, the model is split into two parts: feature extractor $f_\theta$ and classifier $g_\phi$. $f_\theta$ contains a series of convolutional layers followed by fully connected layers which map the image from a higher dimension $\mathbb{R}^{x}$ to class-vector dimension $\mathbb{R}^{h}$. $g_\phi$ is also a classifier (neural network) which maps from $\mathbb{R}^{h}$ to number of classes. We now define our semantic loss in this case as follows:
\begin{equation}
\mathcal{L}^{Semantic}(\theta) = ||(f_\theta({X}_{ij}) -w[Y_{ij}] )||^{2}
\label{eqn:semantic_loss}
\end{equation}
where $w[Y_{ij}]$ denotes the word embedding of the label of image $X_{ij}$.
We modify the aggregation based method to include a semantic loss in a low dimension feature space $\mathbb{R}^{h}$.
\aggzsdg~ is hence trained to minimize the following loss function: 
\begin{equation}
\min_{\theta, \phi} \frac{1}{N}\sum_{i=1}^N \frac{1}{n_i}\sum_{j=1}^{n_i}   \mathcal{L}^{CE}(g_\phi(f_\theta({X}_{ij})), Y_{ij}) + \lambda \cdot \mathcal{L}^{Semantic}(\theta) + \eta\mathcal{R}(\theta,\phi)
\label{eqn:s-agg:total_loss}
\end{equation}
where $\mathcal{L}^{CE}$ is cross-entropy loss, $\lambda$ is weighing factor and $\mathcal{R}$ is a regularizer.\\

\vspace{-22pt}
\subsubsection{\mtaezsdg}
% \noindent \textbf{\mtaezsdg:}
Multi-task Auto Encoder (MTAE)~\cite{ghifary2015domain} learns domain-invariant features using an autoencoding framework. In MTAE, the encoder acts as a feature extractor. A single encoder ($h_{\Theta}$) is maintained across domains, which projects each image to a latent representation. Domain-specific decoders ($g_{\Phi_{1}}, g_{\Phi_{2}}, \ldots g_{\Phi_{N}}$) take these representations and regenerate the image back to each domain. This implicitly forces the encoder to learn an unbiased, domain-agnostic projection function. 
The final classification function is learned using these domain-invariant features with an SVM or a simple MLP. We adapt MTAE for Zero-Shot DG by restricting the latent representation to the semantic embedding of the class labels. 

MTAE ensures self-domain reconstruction and inter-domain reconstruction. For every pair of domains $(D_i,D_j)_{1\leq i,j \leq  N }$, images of domain $D_j$ are generated from images of domain $D_i:  (g_{\Phi_{j}}\circ h_{\Theta})(\boldsymbol{{I}})$  where $\boldsymbol{I}$ is an image of class $k$ in domain $D_j$ that is regenerated (decoded) from images of class $k$ in domain $D_i$. 
While doing the above reconstruction the objective is the features captured by the feature extractor are domain invariant and can be used further for classification.
In standard MTAE we only have a reconstruction loss term. 
Our semantic loss restricts the feature space of MTAE to the semantic embedding space of classes, along with standard reconstruction loss. By doing so, these features also act as a criteria for classification i.e we do not need additional neural network or SVM for classification. 
The loss function for S-MTAE while training domain $i$ is hence: 
\begin{equation}
\mathcal{L}(\Theta, \Phi_1, \ldots, \Phi_N, i) = \sum_{j=1}^{N} \mathcal{L}^{MSE}(g_{\Phi_{j}}(h_{\Theta}(D_i)), D_j) + \lambda \cdot \mathcal{L}^{Semantic}(h_{\Theta}(D_i), w[Y_{D_i}])
\label{eqn:s-mtae_loss}
\end{equation}
where $w[Y_{D_i}]$ is the word embedding of classes in $Y_{D_i}$.
We minimize the following objective to train across $N$ domains: 

\begin{equation}
\min_{\Theta, \Phi_1, \ldots, \Phi_N} \frac{1}{N}\sum_{i=1}^N \mathcal{L}(\Theta, \Phi_1, \ldots, \Phi_N, i ) + \eta\mathcal{R}(\Theta,\Phi_1, \ldots, \Phi_N)
\label{eqn:s-mtae_objective}
\end{equation}

\vspace{-7pt}
\subsubsection{\ourmethod}
% \noindent \textbf{\ourmethod:}
Feature Critic Networks (FC)~\cite{li2019feature} provides a meta-learning approach to DG. It learns a domain-invariant feature extractor by meta-learning an auxiliary loss that `criticizes' the effectiveness of features generated by the feature extractor, when dealing with an unseen domain. The network is trained by simulating training-to-testing domain shift by splitting the source domains into virtual training and testing meta-domains, following standard meta-learning practice.
The goal is to train a model which can perform well both across new domains and also across zero-shot classes. We achieve both these objectives in a novel way i.e model learns to generalize horizontally(across domains) by simulating domain shift and generalize vertically(across zero-shot classes) by exploiting the semantic information of classes. 

Continuing with our generic architecture of domains generalization methods, 
FC can also be viewed as a composition of a feature extractor $f_\theta$ and classifier $g_\phi$.
% Feature extractor($f_\Theta$) maps the image from the higher dimension of image $\mathbb{R}^{x}$ to word vector dimension $\mathbb{R}^{h}$. Classifier($g_\phi$) maps from $\mathbb{R}^{h}$ to number of classes.
% Along with cross-entropy(CE) loss of classifier we introduce a new semantic loss in feature-extractor which restricts the output of the feature-extractor to word vector of the class by taking mean-square-distance between word vector of class and the features extracted by $f_\Theta$.
While training, we split $D$ into $D^{tr}_{meta}, D^{ts}_{meta}$ such that $D^{tr}_{meta} \cap D^{ts}_{meta} = \emptyset$. We train the model on $D^{tr}_{meta}$ but expect it to perform well on $D^{ts}_{meta}$.
The model is trained using the following loss function:
\begin{equation}
\min_{\theta, \phi} \sum_{D_{i} \in D^{tr}_{meta}} \sum_{{X},{Y} \in D_{i}} \mathcal{L}^{CE}(g_{\phi}(h_{\theta}({X})), {Y}) + \lambda \cdot \mathcal{L}^{Semantic}(h_{\theta}({X}), w[{Y}]) + \mathcal{L}^{Aux}
\label{eqn:s-fc_loss}
\end{equation}
$w[{Y}]$ refers to word vectors of classes of ${Y}$, $\mathcal{L}^{CE}$ is cross-entropy loss, and  $\mathcal{L}^{Semantic}$ is the semantic loss defined in Eqn \ref{eqn:semantic_loss}. $\mathcal{L}^{Aux}$ is the meta-loss that encourages feature extractor to generate domain-agnostic features. We refer readers to \cite{li2019feature} for specifics of the meta-training strategy.

\vspace{-10pt}
\section{Experiments and Results}
\label{sec:experiments}

\vspace{-5pt}
\subsection{Datasets}
\label{subSection:datasets}
We evaluate the proposed methods on four different datasets: CIFAR-10 \cite{krizhevsky2009learning}, Fashion-MNIST~\cite{xiao2017fashion}, CIFAR-100 \cite{krizhevsky2009learning} and PACS \cite{li2017deeper}. PACS is the only ready made DG dataset and since PACS contains only 4 domains even its role is limited. We are thus forced to generate new datasets based on rotations. Similar to \cite{deshmukh2019generalization,ghifary2015domain,li2018deep} six different domains are obtained  with $0^{\circ}$ to $75^{\circ}$ rotation and a step of $15^{\circ}$ for each non DG dataset (CIFAR-10, Fashion-MNIST, CIFAR-100). For each non DG dataset $\mathcal{D}$, we have $(\mathcal{D}_{0}, \mathcal{D}_{15}, \mathcal{D}_{30}, \mathcal{D}_{45}, \mathcal{D}_{60}, \mathcal{D}_{75})$ where $\mathcal{D}_{x}$ = images rotated by $x$ degrees. Images are zero-padded as required after rotation. 

PACS\cite{li2017deeper} and Rotated-MNIST\cite{lecun-mnisthandwrittendigit-2010} are often used in earlier DG work \cite{deshmukh2019generalization,ghifary2015domain,li2017learning,li2019feature,li2018deep}. However, we restrain from using Rotated-MNIST because of the lack of connection between the visual space and semantic space of numbers here.

We perform leave-one-out experiments with domains for all datasets. We measure standard DG performance as well as Zero-Shot DG performance on the left-out domain in each experiment. Our experiments are carried out on multiple settings as described below (each setting denotes the zero-shot classes used in the experiment on each dataset):

\noindent (i) \underline{\textit{CIFAR-10:}}
Setting 1: (cat, truck); Setting 2: (cat, dog); Setting 3: (deer, ship); Setting 4: (car, deer); Setting 5: (airplane, car); Same zero-shot classes were used in \cite{socher2013zero}. \\
(ii) \underline{\textit{Fashion-MNIST:}}
% FASHION-MNIST contains 60,000 images and 10 classes : 'Tshirt','Trouser','Pullover','Dress','Coat','Sandal','Shirt',\\
% 'Sneaker','Bag' and 'Boot'. Dataset and implementation details of Rotations of Fashion MNIST is exactly same as Rotations of CIFAR-10 except for the zero-shot setting's for Fashion MNIST change.
% We perform the experiments on 5 zero-shot settings in each leave-one-out domain experiment:\\
Setting 1: (t-shirt, sandal); Setting 2: (sandal, shirt); Setting 3: (t-shirt, boot); Setting 4: (sandal, Boot). \\
(iii) \underline{\textit{CIFAR-100:}}
% CIFAR-100 contains 50,000 images and 100 classes. Each class has 500 images in the dataset. In Rotations of CIFAR-100 we pick the complete dataset per each Domain and rotate accordingly.
% In each domain we have 50000 images. In Rotations of CIFAR-100 In each Zero-Shot setting we leave out 20 classes as zero-shot classes in train-dataset. The total size of train-dataset is $50000 \times 5 - 500 \times 20 \times 5 = 200000$. In test-set we pick all 50000 images of CIFAR-100, rotate it and split it into domain generalization and zero-shot test-sets.
% We perform the experiments on 2 zero-shot settings in each leave-one-out domain experiment:
Setting 1: (whale, fish, rose, can, orange, lamp, couch, beetle, tiger, skyscraper, mountain, kangaroo, fox, snail, man, snake, squirrel, pine-tree, motorcycle, streetcar);
Setting 2: (seal, shark, poppy, bottle, apple, keyboard, table, caterpillar, lion, bridge, forest, camel, raccoon, crab, girl, dinosaur, rabbit, maple, bicycle, tractor). These classes were randomly selected. Only two settings are considered due to the complexity of the dataset.\\
(iv)\underline{\textit{PACS:}} Setting 1: (horse, house); Setting 2: (dog,house); Setting 3: (giraffe,person); Setting 4: (elephant,house). \\

In each setting, independently, ZSDG performance is measured on data of unseen classes from target domain, and DG performance is measured on data of seen classes from target domain. We considered diff settings to study robustness.
\vspace{-10pt}
\subsection{Implementation Details}
% \noindent \textbf{Implementation Details:}

The partition size of $D^{tr}$ and $D^{ts}$ in Semantic FC is 3:2; i.e three domains are chosen as meta-train and two as meta-test.
We compare the performance across 6 different methods: AGG, S-AGG, FC, S-FC, MTAE, S-MTAE. Here AGG, FC and MTAE, we mean the vanilla method without semantic loss and S-\{AGG, MTAE, FC\} denote their semantic counterpart. The ZSDG accuracies are computed using distance in the semantic space to the class vectors. 
% In these cases, even though we don't use semantic loss during training, the ZSDG accuracies are computed using semantic distances (since that is required for zero-shot classification). 
The vanilla method helps us in understanding the usefulness of adding the semantic loss. The reported results are averaged across 5 seeds. All the codes, videos and other resources will be available at \href{https://github.com/aniketde/ZeroShotDG}{https://github.com/aniketde/ZeroShotDG}.\\

We meticulously designed fair experiments for this new setting. As defined in  Sec \ref{subSection:datasets}, we remove some classes (chosen randomly) from domains used for training, from known DG datasets. Trained ZSDG models are evaluated on these unseen classes in unseen domains. Vanilla DG methods form the baseline (which is fair, considering lack of any other methods at this time). The network, dataset, training strategies in vanilla methods and semantic counterparts are all kept consistent for fair comparison.\\

\noindent \textbf{Choice of Word Vectors:}
% \subsubsection{Choice of word vectors}
Word2Vec\cite{Miller95wordnet:a} uses a two-layer neural net that processes text by vectorizing words. GloVe\cite{pennington2014glove} is an unsupervised learning algorithm which uses word-word co-occurrence statistics from a corpus to obtain word vectors. We use the pre-extracted Word2Vec and GloVe vectors from wikipedia provided by \cite{zeroshotstructuredembed}. 
After comparing with different semantic vectors we found GloVe embeddings \cite{pennington2014glove} to be much more meaningful and hence help in getting good results. \\

% All the experiments are run under 5 different seeds and the reported accuracies are in the form of mean $\pm$ standard-deviation. MTAE on CIFAR-100 is only experiment which is run using single seed because rotations of CIFAR-100 contains 250000 images and therefore MTAE becomes computationally expensive.
\vspace{-15pt}
\subsection{Results}
% \noindent \textbf{Results:} 
Tables \ref{tab:avg_fmnist}, \ref{tab:avg_cifar10}, \ref{tab:avg_cifar100} and \ref{tab:avg_pacs} present Domain Generalization (DG) and Zero-Shot Domain Generalization (ZSDG)
results on F-MNIST~\cite{xiao2017fashion}, CIFAR-10 \cite{krizhevsky2009learning}, CIFAR-100 \cite{krizhevsky2009learning} and PACS \cite{li2017deeper} datasets respectively.
For brevity, we average the accuracy across the domains and present the average accuracy in these tables. The complete results with standard deviations across our multiple runs are in the Supplementary material.
We see in the tables that the proposed semantically consistent adaptations of DG methods perform better both on DG and ZSDG. 
On rotations of F-MNIST, S-MTAE performs better than all other methods with the exception of Setting 4. On rotations of CIFAR-10, S-MTAE and S-FC both perform the best. On rotations of CIFAR-100, S-FC performs better when compared to other methods. From the results, one can hypothesize that for simpler datasets, basic DG methods such as MTAE are sufficient and yield good performance, but when the complexity in the dataset increases (as in CIFAR-100), more complex methods such as FC are required for better performance. For purposes of better understanding, we also present in Table \ref{tab:socher} the ZSL performance on the considered settings on the CIFAR-10 dataset using the method proposed by Socher \etal \cite{socher2013zero}. We note that the comparatively lower numbers in Table \ref{tab:avg_cifar10} is because we only use 4000 images from the different domains of CIFAR-10, which is lower than the original dataset. We performed a Wilcoxon signed rank sum test and found that our results of semantic variants are statistically significant at a p-value of 0.04.

% \begin{table}[]
\begin{wraptable}{r}{0pt}
\centering
\resizebox{0.6\textwidth}{!}{%
\begin{tabular}{|l|c|c|c|c|c|c|c|}
\hline
Target: & Setting 1 & Setting 2 & Setting 3 & Setting 4 & Setting 5 \\ \hline
zero-shot & 93.17 & 53.55 & 86.06 & 88.31 & 67.54 \\ \hline
\end{tabular}
}
\caption{Zero-Shot classification accuracies on CIFAR-10 using our implementation of Socher \etal. \cite{socher2013zero}.}
\label{tab:socher}
\end{wraptable}

PACS is widely used in the Domain Generalization papers.  But PACS contains only four domains and seven classes, nevertheless we have performed some experiments on PACS and the results are in table \ref{tab:avg_pacs}. On PACS S-AGG has the best ZSDG accuracy whereas FC has the best DG accuracy. MTAE \& S-MTAE did not perform as expected and hence we have not reported these results. We hypothesize that since PACS contains too few classes, current methods are not suitable for task at hand.

\begin{table}[]
\centering
\resizebox{0.8\textwidth}{!}{%
\begin{tabular}{lcccccccc}
\hline
TARGET & \multicolumn{2}{c}{Setting 1}& \multicolumn{2}{c}{Setting 2}& \multicolumn{2}{c}{Setting 3}& \multicolumn{2}{c}{Setting 4}\\ \cmidrule{2-9}
& DG & ZSDG & DG & ZSDG & DG & ZSDG & DG & ZSDG \\ \hline
AGG & 67.16 & 61.51 & 70.24 & 51.59 & 67.21 & 57.13 & 60.87 & 53.36 \\ \cmidrule{2-9}
S-AGG & 69.11 & 57.673 & 73.19 & 56.87 & 68.08 & 41.88 & 62.37 & 52.5 \\ \cmidrule{2-9}
MTAE & 18.12 & 73.105 & 18.41 & 70.79 & 17.50 & 79.44 & 17.62 & \textbf{64.26} \\ \cmidrule{2-9}
S-MTAE & \textbf{72.97} & \textbf{92.45} & \textbf{77.29} & \textbf{89.54} & \textbf{72.17} & \textbf{89.00} & \textbf{65.56}& 52.71 \\ \cmidrule{2-9}
FC & 66.17 & 56.61 & 69.03 & 52.33 & 66.14 & 53.18 & 59.18 & 51.41 \\ \cmidrule{2-9}
S-FC & 66.53 & 49.03 & 69.93 & 54.24 & 66.33 & 61.04 & 58.83 & 53.94\\ \hline
\end{tabular}
}
\\
\caption{Domain Generalization (DG) and Zero-Shot Domain Generalization (ZSDG) performance of different domains on Fashion-MNIST dataset.}

\label{tab:avg_fmnist}
\end{table}

\begin{table}[]
% \begin{wraptable}{r}{0pt}
\centering
\resizebox{0.9\textwidth}{!}{%
\begin{tabular}{lcccccccccc}
\hline
TARGET &\multicolumn{2}{c}{Setting 1}& \multicolumn{2}{c}{Setting 2}&\multicolumn{2}{c}{Setting 3}&\multicolumn{2}{c}{Setting 4}&\multicolumn{2}{c}{Setting 5}  \\
\cmidrule{2-11}
& DG             & ZSDG              & DG             & ZSDG             & DG             & ZSDG              & DG             & ZSDG              & DG             & ZSDG             \\ \hline
AGG & 51.58 & 48.94 & 50.85 & 49.87 & 48.79 & 42.63 & 49.42 & 52.40 & 47.93 & 51.21 \\ \cmidrule{2-11}
S-AGG & 48.55 & 79.77 & 48.68 & 53.58 & 46.04 & \textbf{81.75} & 46.26 & 82.59 & 44.94 & 65.86 \\ \cmidrule{2-11}
FC & \textbf{52.18} & {54.82} & 51.87 & 50.19 & \textbf{49.56} & 51.69 & \textbf{49.95} & 45.00 & \textbf{48.37} & 52.40 \\ \cmidrule{2-11}
S-FC & 51.35 & \textbf{81.1} & 50.98 & 55.3 & 48.53 & 77.37 & 49.29 & 81.59 & 47.79 & 71.15 \\ \cmidrule{2-11}
MTAE & 12.14 & 54.55 & 11.74 & 51.19 & 12.67 & 56.55 & 12.41 & 52.91 & 12.66 & 54.62 \\ \cmidrule{2-11}
S-MTAE & 51.92 & 80.12 & \textbf{51.94} & \textbf{55.35} & 49.13 & 79.94 & 49.42 & \textbf{83.2} & 47.95 & \textbf{71.63}\\ \hline
\end{tabular}
}

\caption{Domain Generalization (DG) and Zero-Shot Domain Generalization (ZSDG) performance of different domains on CIFAR-10 dataset.}
\label{tab:avg_cifar10}
\end{table}

\begin{table}[!htb]
\begin{minipage}{.34\linewidth}
\resizebox{\textwidth}{!}{%
\begin{tabular}{lcccc}
\hline
&\multicolumn{2}{c}{Setting 1}& \multicolumn{2}{c}{Setting 2}\\
\cmidrule{2-5}
& DG & ZSDG & DG & ZSDG \\
\cmidrule{2-5}
AGG & 80.31 & 5.87 & 80.47 & 6.08 \\ \cmidrule{2-5}
S-AGG & 74.98 & 19.99 & 75.5 & 20.11 \\ \cmidrule{2-5}
FC & \textbf{83.62} & 5.5 & \textbf{83.62} & 5.52 \\ \cmidrule{2-5}
S-FC & 83.47 & \textbf{20.17} & \textbf{83.62} & \textbf{20.7} \\ \cmidrule{2-5}
MTAE & 1.45 & 5.00 & 1.29 & 5.44 \\ \cmidrule{2-5}
S-MTAE & 82.03 & 19.26 & 82.16 & 19.24 \\ \hline
\end{tabular}}
\caption{Domain Generalization (DG) and Zero-Shot Domain Generalization (ZSDG) performance of different domains on CIFAR-100 dataset.}
\label{tab:avg_cifar100}
\end{minipage}\hfill
\begin{minipage}{0.64\linewidth}
    \centering
\resizebox{\textwidth}{!}{%
\begin{tabular}{lcccccccc}
\hline
&\multicolumn{2}{c}{Setting 1}& \multicolumn{2}{c}{Setting 2}& \multicolumn{2}{c}{Setting 3}& \multicolumn{2}{c}{Setting 4}\\
\cmidrule{2-9}
& DG & ZSDG & DG & ZSDG & DG & ZSDG & DG & ZSDG\\
\cmidrule{2-9}
AGG & \textbf{74.33} & 48.03 & 77.39 & 50.9 & 75.48 & 55.47 & 71.64 & 44.06 \\ \cmidrule{2-9}
S-AGG & 69.96 & \textbf{80.13} & 72.125 & \textbf{77.94} & 71.24 & 50.21 & 68.2 & \textbf{66.84} \\ \cmidrule{2-9}
FC & 72.93 & 45.53 & \textbf{77.76} & 46.39 & \textbf{75.68} & 57.37 & \textbf{71.72} & 43.59 \\ \cmidrule{2-9}
S-FC & 66.11 & 76.22 & 67.55 & 74.97 & 67.6 & \textbf{58.45} & 64.03 & 63.08 \\
\hline
\end{tabular}
}
\caption{Domain Generalization (DG) and Zero-Shot Domain Generalization (ZSDG) performance on PACS dataset.}
\label{tab:avg_pacs}
\end{minipage}
% \vspace{-27pt}
\end{table}

\vspace{-15pt}
\section{Ablation Studies}
\vspace{-5pt}
\subsection{Changing Weighting Co-efficients}
A very direct ablation study is changing the weighing factor of the semantic loss and observing the performances. Consider the below general loss function, where $\lambda$ is the weighing factor. \\
\begin{equation}
\min_{\theta} \mathcal{L}^{model}(\theta) + \lambda \cdot \mathcal{L}^{Semantic}(\theta)
\label{eqn:ablation}
\vspace{-5pt}
\end{equation}

Different values for $\lambda$ have been considered and accuracies are averaged over five runs and results are in the Table \ref{tab:ablation}. We infer that as the weight of the semantic loss increases the performance of the model increases. In case of S-MTAE since it is a simpler method the performance goes down when $\lambda$ is very high and optimal performance is observed when $\lambda$ is 5. In case of S-FC and S-AGG the performance improves when the weighing factor increases from 5 to 10 also.

\begin{table}[]
\centering
\resizebox{\textwidth}{!}{%
\begin{tabular}{lccccccccccc}
\hline
& $\lambda$ & \multicolumn{2}{c}{Setting 1}& \multicolumn{2}{c}{Setting 2}& \multicolumn{2}{c}{Setting 3}& \multicolumn{2}{c}{Setting 4}& \multicolumn{2}{c}{Setting 5}\\
\hline
% \cmidrule{2-12}
&  & DG & ZSDG & DG & ZSDG & DG & ZSDG & DG & ZSDG & DG & ZSDG \\
\hline

& 0.1 & 57.42 & \textbf{83.44} & 56.87 & 56.18 & 53.91 & 80.34 & 54.39 & 86.46 & 52.75 & 73.61 \\
& 0.5 & 57.84 & 82.39 & 57.41 & 56.15 & 54.27 & 80.20 & 54.88 & 86.62 & 52.53 & 73.83 \\
S-AGG & 1 & 51.92 & 82.42 & 52.40 & 53.67 & 49.24 & \textbf{84.20} & 49.47 & 84.75 & 47.78 & 67.67 \\
& 5 & 59.04 & 81.69 & 58.62 & 56.30 & 55.66 & 80.25 & 56.46 & 87.48 & 53.89 & 75.34 \\ 
& 10 & \textbf{60.09} & 82.45 & \textbf{59.68} & \textbf{56.36} & \textbf{56.48} & 80.26 & \textbf{57.54} & \textbf{87.62} & \textbf{54.59} & \textbf{75.36} \\ 
\hline
%\cmidrule{2-12}
% &  &  &  &  &  &  &  &  &  &  &  \\ \cmidrule{2-12}
& 0.1 & 53.79 & 82.39 & 53.51 & 54.85 & 51.21 & 77.5 & 50.79 & 82.53 & 49.64 & 70.19 \\
& 0.5 & 54.25 & \textbf{83.69} & 53.56 & 55.29 & 51.30 & 79.84 & 51.18 & 83.93 & 49.72 & 72.42 \\
S-FC & 1 & 54.27 & 82.92 & 54.27 & 55.63 & 51.80 & \textbf{81.57} & \textbf{51.59} & 83.86 & 50.08 & 71.83 \\
& 5 & 54.92 & 82.4 & 54.83 & 56.24 & 52.18 & 80.94 & \textbf{51.59} & 84.00 & 49.99 & 72.58 \\
& 10 & \textbf{55.09} & 82.99 & \textbf{55.79} & \textbf{56.63} & \textbf{52.69} & 80.47 & 51.49 & \textbf{85.59} & \textbf{50.11} & \textbf{74.51} \\ 
\hline
% \cmidrule{-12}
%  &  &  &  &  &  &  &  &  &  &  &  \\ \cmidrule{2-12}
& 0.1 & 56.68 & 83.14 & 55.76 & 56.44 & 52.17 & 81.74 & 52.99 & 86.83 & 49.83 & 75.49 \\
& 0.5 & 54.24 & 79.52 & 54.36 & 56.83 & 52.16 & 81.41 & 52.08 & 84.52 & 48.03 & 71.84 \\
S-MTAE & 1 & 57.22 & 82.03 & 56.36 & 56.34 & 52.70 & 81.26 & 53.17 & 84.85 & 50.77 & 71.98 \\
& 5 & \textbf{58.98} & 82.39 & 58.38 & 56.58 & \textbf{55.82} & \textbf{83.68} & \textbf{55.59} & \textbf{86.86} & 53.27 & \textbf{73.60} \\
& 10 & 58.09 & \textbf{82.74} & \textbf{58.55} & \textbf{56.90} & 55.50 & 83.65 & 56.30 & 87.1 & \textbf{53.78} & 73.38 \\  \hline

\end{tabular}}
\caption{CIFAR-10 Dataset; left-out-domain = $D_3$; Domain Generalization and Zero-Shot Domain Generalization (ZSDG) performances when the weighing factor of the semantic loss is changed.}

\label{tab:ablation}
\end{table}

\subsection{t-SNE Visualization of Semantic Space}

\begin{figure}
\begin{minipage}{0.24\textwidth}
    \includegraphics[width=\textwidth]{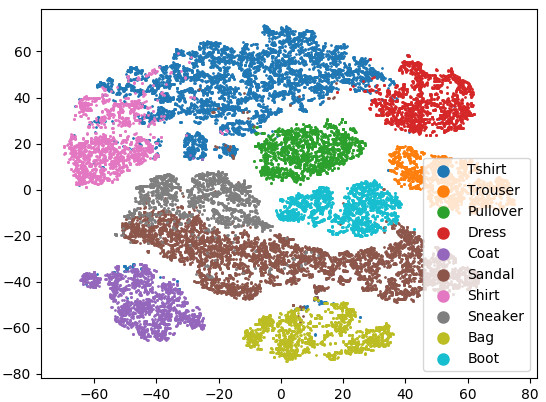}
  \end{minipage}
\begin{minipage}{0.24\textwidth}
    \includegraphics[width=\textwidth]{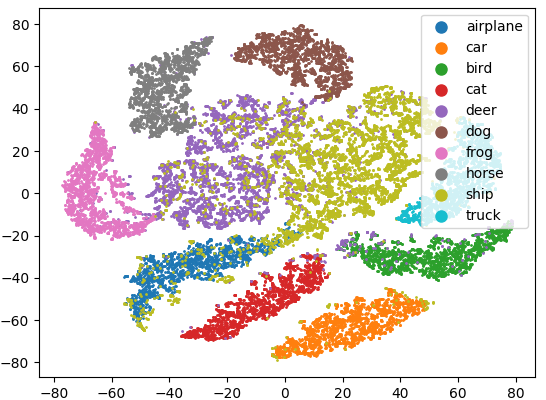}
\end{minipage}
\begin{minipage}{0.24\textwidth}
    \includegraphics[width=\textwidth]{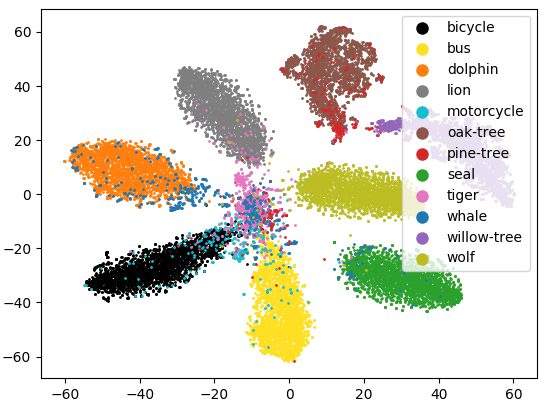}
\end{minipage}
\begin{minipage}{0.24\textwidth}
    \includegraphics[width=\textwidth]{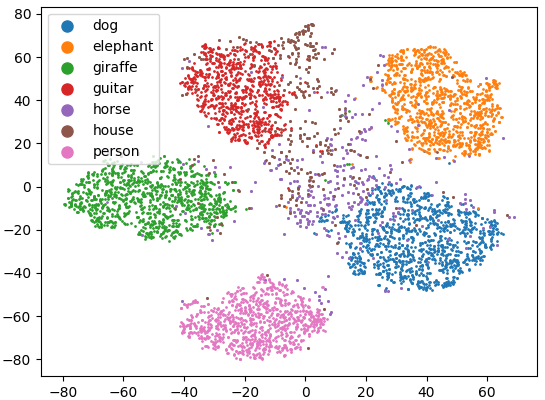}
\end{minipage}
\caption{t-SNE plots of the latent space of ZSDG methods on Fashion-MNIST, CIFAR10, CIFAR100 and PACS datasets respectively (from left to right).}
\label{fig:tsne}
\vspace{-10pt}
\end{figure}

We visualise the semantic space that is learned by the domain generalization methods which has been adapted to solve Zero-Shot Domain Generalization, using our proposed methodology. We use t-SNE \cite{maaten2008visualizing} to project the latent space of the ZSDG methods to two dimension. Figure \ref{fig:tsne} shows the semantic space visualization of models trained on Fashion-MNIST, CIFAR10, CIFAR100 and PACS datasets. Both zero-shot and other classes are plotted in these figures. For fair analysis of different methods, we choose S-MTAE and S-FC for experiments with F-MNIST and PACS dataset, while semantic space of S-AGG is visualised of CIFAR10 and CIFAR100 datasets. 

The domains and the zero-shot classes that are selected are as follows:
\underline{Plot 1}: F-MNIST trained using S-MTAE; target domain: $D_3$; zero-shot classes: t-shirt, sandal.
\underline{Plot 2}: CIFAR10 trained using S-AGG; target domain: $D_2$; zero-shot classes: deer, ship.
\underline{Plot 3}: CIFAR100 trained using S-AGG; target domain: $D_4$; zero-shot classes: Setting 1. Though CIFAR100 contains 100 classes, we randomly sample three zero-shot classes and nine seen classes for the t-SNE plot, for easy visualization.
\underline{Plot 4}: PACS trained using S-FC; target domain: cartoon; zero-shot classes: horse, house.

% The semantic space of the zero-shot images and training dataset images is projected down to two dimensions using t-SNE. Figure \ref{fig:tsne} shows t-SNE visualization of the semantic space of Fashion-MNIST, CIFAR10, CIFAR100 and PACS datasets respectively from left to right.

From Figure \ref{fig:tsne}, we observe that images from the zero shot classes are clustered around semantically meaningful classes. In the first plot, the zero-shot class \textit{t-shirt} is closer to shirt, dress, while the other zero-shot class \textit{sandal} is close to sneaker, boot.
In the second plot, the zero-shot classes \textit{ship} and \textit{deer} is closer to (truck, airplane) and (dog, horse, frog) respectively. Similar observation is found for the other two datasets too. These results allow us to conclude that the semantic loss enforces alignment in the latent space. This enables graceful transition of DG methodologies to solve ZSDG task.

% We can observe that in $\nth{1}$ plot(read from L to R) of \ref{fig:tsne}, the images of zero-shot images of t-shirt are closer to (shirt, dress) whereas zero-shot images of sandal is closer to (sneaker, boot). In \nth{2} the zero-shot images of ship is closer to other transportation vehicles such as (truck, airplane) whereas zero-shot images of deer is closer to other animals such as (dog, horse, frog). Similar observations are found in \nth{3}, \nth{4} as well.
% To-do:
% We can claim that the similarity of word vector has been transferred to the images.

\vspace{-5pt}
\section{Conclusion}
\label{sec:conclusion}

We introduce a novel Zero-Shot Domain Generalization (ZSDG) problem, where a model is expected to generalize to new classes in an unseen domain. This realistic problem setting is harder than Domain Generalization \cite{deshmukh2017multiclass,ghifary2015domain,li2019feature,motiian2017unified}, Domain Adaptation \cite{pan2010domain,daume2009frustratingly,ganin2014unsupervised} and Zero-Shot Learning \cite{socher2013zero,DEVISE_NIPS2013_5204,deepzerotexticcv} as ZSDG models are expected to generalize over novel classes in novel domains, without access to corresponding training data-points. 
We find that learning semantically consistent domain-invariant features helps address this challenging problem. We adapt current state-of-the-art DG methods~\cite{ghifary2015domain,li2019feature} to this setting to reveal the efficacy of our proposed approach, as well as provide a baseline for further efforts on this problem. 

\section*{Acknowledgements}
We thank the anonymous reviewers for their valuable comments and suggestions. This work was partly supported by DST, Govt of India, through the IMPRINT program.
% \clearpage\mbox{}Page \thepage\ of the manuscript.
% \clearpage
% \mbox{}Page \thepage\ of the manuscript.

\bibliography{egbib}

\clearpage
\appendix

\section{Supplementary Section}
% \text{\Large\textbf{Supplementary Section}}\\
% In the supplementary section we will be describing the Preliminaries which was not possible in main paper due to space constraints. We will also be discussing complete results and few observations.
Here, we include the following which could not be included in the main paper due to space limitation:
\begin{itemize}
    \item Preliminary information which would help better understanding of the context.
    \item Complete result, which contains the accuracies of each method on all domains.
\end{itemize}

\subsection{Preliminaries}
\subsubsection{Domain Generalization}
In domain generalization we are given training data from different domains and the objective is to generalize over a novel domain. Let the training dataset be $ D_i = \{X_{ij}, Y_{ij}\}_{1 \leq j \leq n_i}$ and the test dataset be $D_t = \{X_{j}^t, Y_j^t\}_{1 \leq j \leq n_t}$ where $n_{i}$ refers to the number of examples in  $i^{th}$ domain. In domain generalization framework the label space of all $D_i$ and $D_t$ is same. 

We define a domain as the joint distribution of feature and label space such that  \((X_{ij},Y_{ij}) \sim P_{XY}^i\) and \(P_{XY}^i \sim \mu\). The test data set is $(X_j^T, Y_j^T) \sim P^T$ and \(P^T \sim \mu\). We assume all $(X,Y)$ pairs are drawn iid from their respective distributions, and that $P_1, \ldots, P_N, P^T$ are iid from $\mu$.
The main objective of domain generalization is to train a model on all the training domains $D = \{D_1, D_2, \ldots D_N \}$ which performs well on $D^T$.

\subsubsection{Zero Shot Learning}
In zero-shot learning the classes seen during training is disjoint from the classes present in the test phase.
Let $Y^{tr}$ and $Y^{ts}$ represent the set of classes in the training and testing phase.
Let the training dataset be $D^{tr} = \{x_i, y_i\}_{1\leq i\leq n_{tr}  }$, $y_i \in Y^{tr}$. The test dataset be $D^{ts} = \{x_i, y_i\}_{1\leq i\leq n_{ts}}$, $y_i \in Y^{ts}$.
In zero-shot learning framework the objective is to learn a model on $D^{tr}$ which can generalize well on novel classes from $D^{ts}$.

\section{Complete Results}
Section 4.3 in the main paper reports the average accuracy across different domains for each of the method. In this section, we show the complete breakdown of the accuracies across domains.
% Continuing from the discussion in Section 4.3, we now show complete results of our methods on Fashion-MNIST, CIFAR10, CIFAR-100 and PACS. 
Table \ref{tab:full_fmnist}, \ref{tab:full_cifar_10}, \ref{tab:full_cifar_100}, \ref{tab:full_pacs} showcases the domain generalization (DG) results and zero-shot domain generalization (ZSDG)
results on Fashion-MNIST~\cite{xiao2017fashion}, CIFAR-10 \cite{krizhevsky2009learning}, CIFAR-100 \cite{krizhevsky2009learning} and PACS \cite{li2017deeper} datasets. 
All the experiments are run with five different seeds and the reported accuracies are in the form of mean $\pm$ standard-deviation. 
Experiments with MTAE on CIFAR-100 was run using single seed because rotations of CIFAR-100 contains 250,000 images, which made it computationally expensive.
Experiments with large scale datasets like CIFAR 100 provide further insights into the problem.

We observe that semantic loss not only helps with zero-shot domain generalization, but also helps in effectively solving the domain generalization problem as well. This is evident in Table \ref{tab:full_fmnist} where S-MTAE dominates all the other algorithms in DG setting.
Overall, we note that semantic counterparts of the domain generalization methods are able to perform better over the vanilla DG methods while solving zero shot domain generalization problem.

% On rotations of Fashion-MNIST, S-MTAE performs better than all the other methods with the exception of setting 4. On rotations of CIFAR-10 S-MTAE and S-FC both perform better. On rotations of CIFAR-100 S-FC performs better when compared to other methods. From the results we can hypothesize that for easier datasets such as Fashion-MNIST basic domain generalization methods such as MTAE are sufficient and yield good accuracies but when the complexity in the dataset increase we need a complex methods such as Feature-Critic for better performance.

\begin{table}[h]

\centering
\resizebox{\textwidth}{!}{%
\begin{tabular}{llcccccccccc}
\hline
& TARGET &\multicolumn{2}{c}{Setting 1}& \multicolumn{2}{c}{Setting 2}& \multicolumn{2}{c}{Setting 3}& \multicolumn{2}{c}{Setting 4}\\
\cmidrule{3-10}
 &  & DG & ZSDG & DG & ZSDG & DG & ZSDG & DG & ZSDG \\ \cmidrule{3-10}

%  &  &  & 0,5 &  & 5,6 &  & 0,9 &  & 5,9 \\ \hline
AGG &  & 55.33$\pm$1.60 & 65.60$\pm$16.03 & 53.96$\pm$1.40 & 49.08$\pm$2.83 & 55.74$\pm$0.31 & 60.80$\pm$24.64 & \textbf{48.63$\pm$1.41} & 49.11$\pm$12.21 \\ \cmidrule{3-10}
S-AGG &  & 55.11$\pm$2.30 & 57.32$\pm$4.17 & 56.95$\pm$3.67 & 53.86$\pm$3.20 & \textbf{55.93$\pm$2.69} & 62.81$\pm$11.30 & 47.97$\pm$2.48 & 54.15$\pm$4.90 \\ \cmidrule{3-10}
FC & $D_0$ & 53.18$\pm$4.42 & 53.40$\pm$6.48 & 54.35$\pm$3.99 & 54.18$\pm$9.23 & 54.55$\pm$1.33 & 54.81$\pm$11.35 & 46.68$\pm$1.21 & 50.04$\pm$1.23 \\ \cmidrule{3-10}
S-FC &  & 54.52$\pm$2.17 & 54.52$\pm$2.17 & 56.41$\pm$1.23 & 50.23$\pm$3.93 & 55.52$\pm$2.71 & 59.71$\pm$5.23 & 46.38$\pm$2.84 & 51.20$\pm$2.38 \\ \cmidrule{3-10}
MTAE &  & 16.85$\pm$5.75 & 71.18$\pm$10.87 & 21.22$\pm$5.28 & \textbf{79.10$\pm$11.76} & 19.46$\pm$4.93 & \textbf{81.61$\pm$18.32} & 18.48$\pm$6.44 & 60.88$\pm$12.21 \\ \cmidrule{3-10}
S-MTAE &  & \textbf{55.82$\pm$4.40} & \textbf{78.95$\pm$2.65} & \textbf{57.97$\pm$5.02} & 70.61$\pm$3.10 & 52.20$\pm$2.75 & 64.63$\pm$8.29 & 44.41$\pm$3.90 & \textbf{63.56$\pm$8.53} \\ \cmidrule{3-10}

 &  &  &  &  &  &  &  &  &  \\ \cmidrule{3-10}
AGG &  & 75.87$\pm$0.94 & 58.70$\pm$11.95 & 81.58$\pm$1.65 & 52.25$\pm$8.01 & 75.98$\pm$1.06 & 57.85$\pm$18.11 & 69.95$\pm$1.96 & 55.23$\pm$8.34 \\ \cmidrule{3-10}
S-AGG &  & 76.57$\pm$0.88 & 67.87$\pm$3.94 & 82.37$\pm$0.76 & 57.41$\pm$2.01 & 76.64$\pm$0.40 & 35.78$\pm$6.05 & 70.70$\pm$1.88 & \textbf{57.70$\pm$8.04} \\ \cmidrule{3-10}
FC & $D_1$ & 75.26$\pm$1.22 & 47.96$\pm$15.50 & 79.87$\pm$1.58 & 36.65$\pm$14.88 & 74.75$\pm$1.10 & 49.15$\pm$1.86 & 68.12$\pm$2.42 & 52.46$\pm$1.76 \\ \cmidrule{3-10}
S-FC &  & 74.41$\pm$1.39 & 30.46$\pm$8.63 & 79.14$\pm$2.36 & 53.47$\pm$7.97 & 73.4$\pm$1.43 & 41.44$\pm$9.10 & 66.66$\pm$2.99 & 51.01$\pm$1.62 \\ \cmidrule{3-10}
MTAE &  & 19.11$\pm$6.2 & 73.92$\pm$13.76 & 20.36$\pm$3.85 & 71.35$\pm$3.46 & 14.69$\pm$5.44 & 83.62$\pm$15.65 & 17.84$\pm$3.41 & 57.50$\pm$14.93 \\ \cmidrule{3-10}
S-MTAE &  & \textbf{79.43$\pm$1.31} & \textbf{91.51$\pm$2.38} & \textbf{84.67$\pm$2.44} & \textbf{87.32$\pm$1.12} & \textbf{79.09$\pm$0.86} & \textbf{90.64$\pm$8.78} & \textbf{73.73$\pm$2.09} & 55$\pm$4.27 \\ \cmidrule{3-10}

 &  &  &  &  &  &  &  &  &  \\ \cmidrule{3-10}
AGG &  & 74.06$\pm$1.42 & 62.23$\pm$18.68 & 79.23$\pm$1.78 & 52.03$\pm$15.65 & 73.97$\pm$1.07 & 52.69$\pm$21.74 & 68.35$\pm$1.29 & 50.67$\pm$6.02 \\ \cmidrule{3-10}
S-AGG & & 75.46$\pm$1.00 & 51.89$\pm$2.44 & 80.54$\pm$1.27 & 14.03$\pm$2.14 & 75.34$\pm$2.60 & 10.58$\pm$2.60 & 68.65$\pm$0.63 & 55.78$\pm$8.18 \\ \cmidrule{3-10}
FC & $D_2$ & 73.42$\pm$1.36 & 69.62$\pm$17.44 & 77.67$\pm$2.50 & 58.69$\pm$9.41 & 73.50$\pm$1.87 & 61.98$\pm$13.79 & 67.36$\pm$1.57 & 52.45$\pm$3.40 \\ \cmidrule{3-10}
S-FC &  & 73.33$\pm$1.09 & 40.94$\pm$5.97 & 77.02$\pm$1.58 & 54.37$\pm$3.00 & 73.68$\pm$0.85 & 55.13$\pm$4.77 & 65.4$\pm$0.97 & \textbf{53.99$\pm$3.98} \\ \cmidrule{3-10}
MTAE &  & 20.73$\pm$3.52 & 65.57$\pm$18.23 & 16.97$\pm$3.41 & 66.21$\pm$17.80 & 15.12$\pm$7.77 & 75.52$\pm$19.55 & 15.13$\pm$3.54 & 69.38$\pm$10.73 \\ \cmidrule{3-10}
S-MTAE &  & \textbf{78.08$\pm$0.69} & \textbf{94.26$\pm$1.03} & \textbf{83.39$\pm$1.28} & \textbf{92.03$\pm$1.52} & \textbf{77.75$\pm$1.07} & \textbf{92.54$\pm$4.72} & \textbf{70.74$\pm$0.61} & 50.10$\pm$3.81 \\ \cmidrule{3-10}

 &  &  &  &  &  &  &  &  &  \\ \cmidrule{3-10}
AGG &  & 72.70$\pm$1.37 & 62.52$\pm$26.81 & 77.13$\pm$1.81 & 47.46$\pm$27.99 & 72.75$\pm$0.80 & 57.99$\pm$21.65 & 67.08$\pm$0.80 & 52.72$\pm$5.35 \\ \cmidrule{3-10}
S-AGG &  & 74.84$\pm$1.57 & 53.65$\pm$2.30 & 80.04$\pm$1.58 & 70.91$\pm$3.61 & 75.11$\pm$0.66 & 49.18$\pm$0.49 & 68.16$\pm$1.55 & 52.22$\pm$4.05 \\ \cmidrule{3-10}
FC & $D_3$ & 72.29$\pm$1.17 & 59.45$\pm$23.74 & 76.52$\pm$2.10 & 53.58$\pm$10.01 & 71.84$\pm$1.32 & 51.45$\pm$24.36 & 65.44$\pm$0.41 & 49.43$\pm$4.49 \\ \cmidrule{3-10}
S-FC &  & 72.87$\pm$1.21 & 67.54$\pm$8.12 & 76.68$\pm$2.56 & 83.23$\pm$9.01 & 71.95$\pm$1.02 & 82.22$\pm$13.5 & 64.90$\pm$1.34 & 51.87$\pm$2.37 \\ \cmidrule{3-10}
MTAE &  & 20.92$\pm$14.23 & 77.11$\pm$8.28 & 15.56$\pm$4.61 & 69.80$\pm$17.75 & 16.77$\pm$4.25 & 73.05$\pm$17.10 & 21.18$\pm$2.46 & \textbf{61.04$\pm$6.14} \\ \cmidrule{3-10}
S-MTAE &  & \textbf{77.60$\pm$1.50} & \textbf{96.4$\pm$0.63} & \textbf{83.54$\pm$1.48} & \textbf{94.83$\pm$2.75} & \textbf{77.42$\pm$0.43} & \textbf{91.03$\pm$4.94} & \textbf{72.76$\pm$2.51} & 49.17$\pm$0.53 \\ \cmidrule{3-10}

 &  &  &  &  &  &  &  &  &  \\ \cmidrule{3-10}
AGG &  & 75.35$\pm$1.19 & 52.45$\pm$25.22 & 80.36$\pm$1.41 & 50.65$\pm$9.08 & 75.57$\pm$1.02 & 48.77$\pm$33.40 & 70.25$\pm$1.11 & 53.58$\pm$4.31 \\ \cmidrule{3-10}
S-AGG &  & 76.54$\pm$1.20 & 51.62$\pm$1.68 & 82.67$\pm$1.07 & 72.04$\pm$3.86 & 77.18$\pm$0.72 & 49.68$\pm$0.39 & 70.50$\pm$1.86 & 51.91$\pm$5.99 \\ \cmidrule{3-10}
FC & $D_4$ & 75.23$\pm$0.81 & 54.57$\pm$18.00 & 80.06$\pm$1.92 & 52.30$\pm$10.47 & 75.07$\pm$0.83 & 53.87$\pm$17.51 & 68.73$\pm$1.06 & 54.87$\pm$6.90 \\ \cmidrule{3-10}
S-FC &  & 73.77$\pm$0.79 & 50.01$\pm$0.02 & 79.36$\pm$1.99 & 32.83$\pm$9.54 & 73.83$\pm$1.67 & 46.10$\pm$3.23 & 67.44$\pm$0.79 & 65.57$\pm$6.44 \\ \cmidrule{3-10}
MTAE &  & 16.37$\pm$1.81 & 67.55$\pm$12.57 & 19.67$\pm$3.21 & 74.76$\pm$14.21 & 19.43$\pm$2.11 & 75.15$\pm$15.29 & 15.27$\pm$4.99 & \textbf{68.26$\pm$7.61} \\ \cmidrule{3-10}
S-MTAE &  & \textbf{80.99$\pm$1.13} & \textbf{97.97$\pm$0.53} & \textbf{86.22$\pm$1.14} & \textbf{97.84$\pm$0.53} & \textbf{80.10$\pm$0.98} & \textbf{97.24$\pm$2.35} & \textbf{75.62$\pm$1.21} & 49.85$\pm$0.53 \\ \cmidrule{3-10}

 &  &  &  &  &  &  &  &  &  \\ \cmidrule{3-10}
AGG &  & 49.65$\pm$2.56 & 67.60$\pm$14.64 & 49.18$\pm$2.57 & 58.12$\pm$14.59 & 49.27$\pm$2.24 & 64.71$\pm$18.83 & 40.97$\pm$2.75 & 58.90$\pm$10.51 \\ \cmidrule{3-10}
S-AGG &  & 56.14$\pm$1.56 & 63.69$\pm$2.54 & 56.57$\pm$2.36 & 72.98$\pm$0.74 & 48.28$\pm$2.80 & 43.25$\pm$6.41 & 48.28$\pm$2.80 & 43.25$\pm$6.41 \\ \cmidrule{3-10}
FC & $D_5$ & 47.69$\pm$4.35 & 54.68$\pm$10.87 & 45.74$\pm$1.85 & 58.63$\pm$12.93 & 47.18$\pm$2.83 & 47.84$\pm$10.82 & 38.78$\pm$2.55 & 49.25$\pm$2.96 \\ \cmidrule{3-10}
S-FC &  & 50.33$\pm$3.03 & 50.74$\pm$0.58 & 50.98$\pm$2.79 & 51.36$\pm$0.67 & 49.61$\pm$3.72 & 81.64$\pm$13.09 & 42.22$\pm$2.60 & 50.02$\pm$0.04 \\ \cmidrule{3-10}
MTAE &  & 14.78$\pm$2.28 & 83.31$\pm$5.91 & 16.76$\pm$2.16 & 63.55$\pm$10.74 & 19.56$\pm$9.67 & 87.73$\pm$4.96 & 17.84$\pm$3.10 & \textbf{68.53$\pm$13.33} \\ \cmidrule{3-10}
S-MTAE &  & \textbf{65.93$\pm$2.13} & \textbf{95.62$\pm$2.47} & \textbf{68.00$\pm$3.62} & \textbf{94.64$\pm$2.72} & \textbf{66.47$\pm$3.03} & \textbf{97.94$\pm$1.24} & \textbf{56.13$\pm$3.67} & 48.63$\pm$1.30 \\ \hline
\end{tabular}
}
\caption{Domain Generalization (DG) and Zero-Shot Domain Generalization (ZSDG) performance of different domains on \textbf{Fashion-MNIST} dataset.}
\label{tab:full_fmnist}
\end{table}
\clearpage

\begin{table}[]
\centering
\resizebox{\textwidth}{!}{%
\begin{tabular}{llcccccccccc}
\hline
 & TARGET &\multicolumn{0}{r}{Setting 1}& \multicolumn{2}{r}{Setting 2}&\multicolumn{2}{r}{Setting 3}&\multicolumn{2}{r}{Setting 4}&\multicolumn{2}{r}{Setting 5} \\
\cmidrule{3-12}
&         & DG             & ZSDG              & DG             & ZSDG             & DG             & ZSDG              & DG             & ZSDG              & DG             & ZSDG             \\
\cmidrule{3-12}

AGG &      & \textbf{43.98$\pm$1.88} & 49.67$\pm$9.49  & \textbf{43.92$\pm$1.05} & 49.45$\pm$1.94 & 41.38$\pm$1.26 & 45.20$\pm$10.70 & 42.13$\pm$0.83 & 51.93$\pm$16.18 & 40.82$\pm$1.09 & 51.68$\pm$6.70 \\
S-AGG  &         & 39.26$\pm$1.28 & 73.28$\pm$2.94  & 40.28$\pm$0.33 & 52.98$\pm$0.87 & 38.38$\pm$0.25 & \textbf{76.77$\pm$1.65}  & 37.58$\pm$1.67 & \textbf{77.26$\pm$1.97}  & 36.92$\pm$1.66 & 60.10$\pm$2.18 \\ 
FC     & $D_0$        & 43.81$\pm$0.83 & 52.20$\pm$11.78 & 43.86$\pm$0.87 & 50.42$\pm$0.95 & \textbf{41.74$\pm$0.84} & 50.66$\pm$3.58  & \textbf{43.32$\pm$0.94} & 45.03$\pm$6.97  & \textbf{41.52$\pm$0.72} & 52.70$\pm$2.72 \\
S-FC   &         & 43.09$\pm$1.20 & \textbf{74.62$\pm$2.70}  & 43.20$\pm$1.28 & \textbf{54.89$\pm$0.97} & 40.41$\pm$1.05 & 67.99$\pm$4.36  & 42.38$\pm$0.84 & 74.95$\pm$2.46  & 40.91$\pm$1.38 & \textbf{68.01$\pm$5.99} \\
MTAE   &         & 11.52$\pm$1.38 & 52.46$\pm$5.01  & 9.83$\pm$5.31  & 52.77$\pm$2.18 & 14.05$\pm$1.75 & 59.63$\pm$11.88 & 12.12$\pm$1.67 & 54.09$\pm$6.50  & 12.35$\pm$1.70 & 53.60$\pm$4.32 \\
S-MTAE &         & 38.08$\pm$4.31 & 73.23$\pm$5.22  & 39.74$\pm$2.91 & 53.51$\pm$1.47 & 36.28$\pm$1.93 & 72.75$\pm$3.32  & 37.05$\pm$1.59 & 75.67$\pm$2.62  & 36.76$\pm$1.27 & 65.94$\pm$1.87 \\ \cmidrule{3-12}

&         &                &                 &                &                &                &                 &                &                 &                &               \\ \cmidrule{3-12}
AGG    & & 54.83$\pm$0.79 & 49.01$\pm$12.77 & 53.79$\pm$0.60 & 49.78$\pm$3.13 & 52.08$\pm$1.03 & 43.26$\pm$14.61 & 52.58$\pm$1.09 & 53.77$\pm$18.67 & 50.63$\pm$1.22 & 50.85$\pm$8.82 \\
S-AGG  &         & 51.36$\pm$1.41 & 82.26$\pm$0.37  & 51.18$\pm$1.67 & 54.13$\pm$1.11 & 48.78$\pm$1.78 & \textbf{84.21$\pm$1.40}  & 48.80$\pm$1.24 & 84.84$\pm$0.57  & 47.66$\pm$1.39 & 67.33$\pm$1.31 \\
FC     & $D_1$        & \textbf{55.85$\pm$0.46} & 54.05$\pm$15.35 & 55.31$\pm$0.60 & 50.16$\pm$0.77 & \textbf{52.80$\pm$0.80} & 52.56$\pm$2.91  & \textbf{53.30$\pm$0.48} & 43.99$\pm$13.70 & \textbf{51.42$\pm$0.90} & 51.53$\pm$3.78 \\
S-FC   &         & 54.29$\pm$0.89 & \textbf{83.17$\pm$1.64}  & 53.70$\pm$0.35 & 55.48$\pm$1.03 & 50.76$\pm$0.84 & 80.18$\pm$3.00  & 52.40$\pm$1.05 & \textbf{84.86$\pm$0.82}  & 50.25$\pm$1.34 & 72.64$\pm$4.03 \\
MTAE   &         & 11.85$\pm$2.78 & 53.10$\pm$6.58  & 9.82$\pm$5.14  & 50.67$\pm$0.74 & 13.31$\pm$1.31 & 60.15$\pm$10.68 & 13.01$\pm$2.43 & 54.84$\pm$4.60  & 14.29$\pm$2.48 & 51.84$\pm$1.48 \\
S-MTAE &         & 55.34$\pm$1.61 & 81.38$\pm$0.91  & \textbf{55.38$\pm$1.04} & \textbf{56.27$\pm$0.62} & 51.76$\pm$1.78 & 81.08$\pm$1.88  & 52.33$\pm$0.73 & 84.28$\pm$1.86  & 50.87$\pm$2.53 & \textbf{73.54$\pm$2.37} \\ \cmidrule{3-12}
&         &                &                 &                &                &                &                 &                &                 &                &                \\ \cmidrule{3-12}
AGG    &      & 55.98$\pm$0.97 & 48.12$\pm$11.63 & 55.31$\pm$1.10 & 49.94$\pm$3.64 & 53.02$\pm$0.51 & 41.50$\pm$11.43 & 53.54$\pm$0.75 & 52.51$\pm$18.52 & 52.40$\pm$0.58 & 50.33$\pm$4.17 \\
S-AGG  &         & 53.52$\pm$1.25 & 82.99$\pm$0.87  & 52.97$\pm$1.15 & 54.04$\pm$1.39 & 50.44$\pm$0.55 & \textbf{84.60$\pm$2.07}  & 51.44$\pm$1.37 & 85.50$\pm$0.78  & 49.58$\pm$1.09 & 68.48$\pm$1.97 \\
FC     & $D_2$        & 56.70$\pm$1.04 & 55.88$\pm$16.21 & 56.03$\pm$0.87 & 50.30$\pm$1.11 & 53.60$\pm$0.88 & 52.32$\pm$5.38  & 53.82$\pm$0.96 & 43.37$\pm$9.78  & \textbf{52.57$\pm$0.24} & 52.23$\pm$2.65 \\
S-FC   &         & 55.86$\pm$0.79 & \textbf{84.77$\pm$1.48}  & 55.37$\pm$0.49 & 55.83$\pm$0.68 & 53.14$\pm$0.67 & 81.77$\pm$2.96  & 53.02$\pm$1.47 & 84.52$\pm$2.35  & 51.91$\pm$1.16 & 72.45$\pm$3.36 \\
MTAE   &         & 10.63$\pm$4.72 & 56.44$\pm$6.23  & 13.72$\pm$1.39 & 50.93$\pm$1.07 & 13.16$\pm$1.66 & 59.11$\pm$11.37 & 12.40$\pm$1.19 & 51.99$\pm$3.54  & 12.33$\pm$4.03 & 55.71$\pm$3.76 \\
S-MTAE &         & \textbf{56.92$\pm$1.41} & 82.82$\pm$1.05  & \textbf{57.00$\pm$0.79} & \textbf{56.34$\pm$0.74} & \textbf{54.01$\pm$1.98} & 84.54$\pm$1.71  & \textbf{54.96$\pm$1.49} & \textbf{87.09$\pm$1.16}  & 52.25$\pm$1.26 & \textbf{72.84$\pm$1.86} \\ \cmidrule{3-12}
      &         &                &                 &                &                &                &                 &                &                 &                &                \\ \cmidrule{3-12}
AGG    &      & 54.92$\pm$0.80 & 49.98$\pm$10.45 & 53.99$\pm$0.80 & 49.98$\pm$3.07 & 51.81$\pm$1.16 & 40.87$\pm$12.03 & 52.27$\pm$0.79 & 51.96$\pm$17.96 & 50.59$\pm$1.17 & 51.41$\pm$5.32 \\
S-AGG  &         & 51.92$\pm$1.00 & 82.42$\pm$1.25  & 52.40$\pm$1.29 & 53.67$\pm$1.36 & 49.24$\pm$0.53 & \textbf{84.20$\pm$1.62}  & 49.47$\pm$1.07 & 84.75$\pm$0.84  & 47.78$\pm$1.14 & 67.67$\pm$1.48 \\
FC     & $D_3$        & 55.55$\pm$0.94 & 56.65$\pm$11.81 & 55.41$\pm$0.83 & 49.88$\pm$0.99 & 52.61$\pm$1.03 & 53.16$\pm$5.25  & 52.41$\pm$1.65 & 43.34$\pm$8.43  & \textbf{50.88$\pm$1.82} & 51.29$\pm$3.34 \\
S-FC   &         & 54.27$\pm$1.82 & \textbf{82.92$\pm$1.69}  & 54.27$\pm$1.34 & 55.63$\pm$2.65 & 51.80$\pm$1.38 & 81.57$\pm$4.28  & 51.59$\pm$0.43 & 83.86$\pm$1.37  & 50.08$\pm$1.35 & 71.83$\pm$2.94 \\
MTAE   &         & 11.66$\pm$4.37 & 58.13$\pm$4.41  & 12.96$\pm$1.42 & 51.29$\pm$1.64 & 12.64$\pm$0.46 & 49.16$\pm$5.73  & 13.04$\pm$1.00 & 51.20$\pm$5.94  & 13.58$\pm$1.43 & 57.38$\pm$5.85 \\
S-MTAE &         & \textbf{57.22$\pm$0.33} & 82.03$\pm$1.77  & \textbf{56.36$\pm$1.19} & \textbf{56.34$\pm$0.64} & \textbf{52.70$\pm$1.06} & 81.26$\pm$2.40  & \textbf{53.17$\pm$1.18} & \textbf{84.85$\pm$1.76}  & 50.77$\pm$1.39 & \textbf{71.98$\pm$2.48} \\ \cmidrule{3-12}
      &         &                &                 &                &                &                &                 &                &                 &                &                \\ \cmidrule{3-12}
AGG    &      & 54.32$\pm$0.56 & 49.99$\pm$9.20  & 52.9$\pm$0.67  & 50.04$\pm$3.09 & 50.80$\pm$0.34 & 40.14$\pm$11.88 & 51.57$\pm$0.57 & 51.97$\pm$18.27 & 50.01$\pm$1.00 & 51.71$\pm$6.28 \\
S-AGG  &         & 51.64$\pm$1.99 & 82.30$\pm$1.0   & 51.5$\pm$1.27  & 53.45$\pm$0.77 & 48.91$\pm$1.55 & 83.47$\pm$2.20  & 48.80$\pm$1.45 & 83.96$\pm$1.14  & 47.63$\pm$1.46 & 67.57$\pm$1.49 \\
FC     & $D_4$        & 54.82$\pm$0.76 & 56.22$\pm$14.31 & 54.34$\pm$0.52 & 49.79$\pm$0.63 & 51.86$\pm$0.92 & 51.36$\pm$4.38  & 52.13$\pm$0.77 & 47.56$\pm$9.94  & 50.47$\pm$0.60 & 52.64$\pm$2.94 \\
S-FC   &         & 54.17$\pm$1.21 & 83.13$\pm$1.47  & 53.41$\pm$0.87 & \textbf{55.38$\pm$0.62} & 50.52$\pm$1.36 & 80.63$\pm$3.26  & 51.75$\pm$1.02 & 84.05$\pm$1.98  & 49.88$\pm$0.79 & 71.55$\pm$3.37 \\
MTAE   &         & 13.78$\pm$3.39 & 53.79$\pm$4.63  & 12.23$\pm$4.09 & 51.14$\pm$1.47 & 11.41$\pm$4.78 & 56.12$\pm$7.81  & 11.1$\pm$2.03  & 53.03$\pm$4.48  & 12.16$\pm$1.35 & 53.85$\pm$3.68 \\
S-MTAE &         & \textbf{56.61$\pm$1.30} & \textbf{83.36$\pm$1.60}  & \textbf{56.03$\pm$1.58} & 55.36$\pm$0.68 & \textbf{53.69$\pm$0.75} & \textbf{84.38$\pm$1.44}  & \textbf{53.96$\pm$0.42} & \textbf{87.09$\pm$1.49}  & \textbf{52.73$\pm$1.06} & \textbf{73.73$\pm$1.30} \\ \cmidrule{3-12}
      &         &                &                 &                &                &                &                 &                &                 &                &                \\ \cmidrule{3-12}
AGG    &      & 45.50$\pm$0.77 & 49.60$\pm$7.48  & 45.21$\pm$1.19 & 50.03$\pm$2.76 & 43.65$\pm$0.73 & 44.83$\pm$8.22  & 44.45$\pm$0.55 & 52.28$\pm$16.93 & 43.17$\pm$0.98 & 51.32$\pm$8.59 \\ 
S-AGG  &         & 43.65$\pm$0.81 & 75.41$\pm$2.98  & 43.75$\pm$1.29 & 53.21$\pm$0.47 & 40.73$\pm$1.50 & \textbf{77.26$\pm$1.29}  & 41.50$\pm$1.21 & 79.25$\pm$1.76  & 40.11$\pm$0.72 & 64.04$\pm$1.73 \\ 
FC     & $D_5$        & 46.40$\pm$0.90 & 53.92$\pm$13.88 & 46.28$\pm$0.99 & 50.61$\pm$1.09 & 44.79$\pm$0.21 & 50.13$\pm$8.69  & 44.76$\pm$0.88 & 46.76$\pm$7.40  & 43.41$\pm$0.89 & 54.06$\pm$5.11 \\ 
S-FC   &         & 46.45$\pm$2.06 & \textbf{78.04$\pm$2.25}  & 45.97$\pm$0.76 & \textbf{54.60$\pm$0.52} & 44.60$\pm$0.97 & 72.09$\pm$4.84  & 44.61$\pm$1.05 & 77.34$\pm$2.49  & 43.74$\pm$0.83 & 70.42$\pm$4.30 \\ 
MTAE   &         & 13.40$\pm$2.15 & 53.41$\pm$4.68  & 11.93$\pm$5.00 & 50.39$\pm$1.20 & 11.49$\pm$2.17 & 55.13$\pm$6.83  & 12.82$\pm$3.41 & 52.35$\pm$2.13  & 11.29$\pm$4.51 & 55.34$\pm$2.07 \\ 
S-MTAE &         & \textbf{47.40$\pm$0.62} & 77.9$\pm$1.99   & \textbf{47.16$\pm$1.15} & 54.31$\pm$0.84 & \textbf{46.39$\pm$1.26} & 75.63$\pm$1.92  & \textbf{45.08$\pm$0.53} & \textbf{80.22$\pm$3.06} & \textbf{44.36$\pm$1.80} & \textbf{71.76$\pm$1.25} \\ 
\hline
\end{tabular}%
}
\caption{Domain Generalization (DG) and Zero-Shot Domain Generalization (ZSDG) performance of different domains on \textbf{CIFAR-10} dataset.}
\label{tab:full_cifar_10}
\end{table}

% \clearpage

\begin{table}[]
\centering
\resizebox{0.5\textwidth}{!}{%
\begin{tabular}{lccccc}
\hline
& TARGET &\multicolumn{2}{c}{Setting 1}& \multicolumn{2}{c}{Setting 2}\\
\cmidrule{3-6}
&  & DG & ZSDG & DG & ZSDG \\
\cmidrule{3-6}
AGG &  & 49.73$\pm$0.26 & 6.07$\pm$0.61 & 49.40$\pm$0.97 & 5.94$\pm$0.91 \\
\cmidrule{3-6}
S-AGG &  & 37.91$\pm$0.70 & 16.03$\pm$0.20 & 39.14$\pm$0.74 & 18.00$\pm$0.17 \\
\cmidrule{3-6}
FC & $D_0$ & \textbf{53.42$\pm$0.57} & 5.68$\pm$1.24 & \textbf{53.76$\pm$0.48} & 5.62$\pm$1.37 \\
\cmidrule{3-6}
S-FC &  & 52.47$\pm$2.67 & \textbf{17.13$\pm$0.95} & 52.46$\pm$2.73 & \textbf{18.64$\pm$0.81} \\
\cmidrule{3-6}
MTAE &  & 1.275 & 5 & 1.25 & 4.9 \\
\cmidrule{3-6}
S-MTAE &  & 26.68 & 15.5 & 27.66 & 15.05 \\
\cmidrule{3-6}
 &  &  &  &  &  \\
 \cmidrule{3-6}

AGG &  & 90.23$\pm$0.13 & 5.97$\pm$0.59 & 90.41$\pm$0.59 & 6.33$\pm$0.54 \\
\cmidrule{3-6}
S-AGG &  & 77.56$\pm$0.94 & 19.77$\pm$0.45 & 77.96$\pm$0.64 & 20.30$\pm$0.08 \\
\cmidrule{3-6}
FC & $D_1$ & 95.25$\pm$0.27 & 5.65$\pm$0.81 & 95.02$\pm$0.31 & 5.32$\pm$1.34 \\
\cmidrule{3-6}
S-FC &  & \textbf{95.53$\pm$1.43} & \textbf{21.62$\pm$2.31} & \textbf{95.53$\pm$1.43} & \textbf{21.6$\pm$2.31} \\
\cmidrule{3-6}
MTAE &  & 1.815 & 5.03 & 1.25 & 5.25 \\
\cmidrule{3-6}
S-MTAE &  & 88.85 & 18.44 & 90.37 & 18.25 \\
\cmidrule{3-6}
 &  &  &  &  &  \\
 \cmidrule{3-6}
AGG &  & 92.77$\pm$0.37 & 6.07$\pm$1.13 & 93.05$\pm$0.80 & 6.27$\pm$0.60 \\
\cmidrule{3-6}
S-AGG &  & 85.00$\pm$0.72 & 20.66$\pm$0.80 & 85.61$\pm$0.72 & 20.58$\pm$0.58 \\
\cmidrule{3-6}
FC & $D_2$ & 96.37$\pm$0.34	& 5.32$\pm$0.53 & 96.40$\pm$0.36 & 5.27$\pm$1.04 \\
\cmidrule{3-6}
S-FC &  & 95.97$\pm$0.63 & \textbf{21.32$\pm$1.83} & \textbf{96.95$\pm$0.69} & \textbf{21.63$\pm$1.10} \\
\cmidrule{3-6}
MTAE &  & 1.395 & 5 & 1.47 & 5.37 \\
\cmidrule{3-6}
S-MTAE &  & \textbf{98.48} & 19.45 & 96.59 & 20.43 \\
\cmidrule{3-6}
 &  &  &  &  &  \\
 \cmidrule{3-6}
AGG &  & 93.21$\pm$0.79 & 5.79$\pm$0.51 & 93.40$\pm$0.76 & 6.12$\pm$0.66 \\
\cmidrule{3-6}
S-AGG &  & 88.61$\pm$1.04 & \textbf{21.07$\pm$0.42} & 88.87$\pm$1.29 & 20.41$\pm$0.62 \\
\cmidrule{3-6}
FC & $D_3$ & 95.74$\pm$0.35	& 5.56$\pm$0.78	& 95.75$\pm$0.47	& 5.27$\pm$1.59 \\
\cmidrule{3-6}
S-FC &  & 96.14$\pm$0.73 & 20.92$\pm$0.72 & 95.94$\pm$1.14 & \textbf{21.73$\pm$1.01} \\
\cmidrule{3-6}
MTAE &  & 1.432 & 5 & 1.25 & 6.47 \\
\cmidrule{3-6}
S-MTAE &  & \textbf{99.72} & 19.78 & \textbf{99.67} & 20.53 \\
\cmidrule{3-6}
 &  &  &  &  &  \\
 \cmidrule{3-6}
AGG & & 94.55$\pm$0.45 & 5.97$\pm$1.08 & 94.89$\pm$0.64 & 6.11$\pm$0.44 \\ \cmidrule{3-6}
S-AGG &  & 93.90$\pm$0.49 & 22.38$\pm$0.50 & 94.09$\pm$0.33 & 21.62$\pm$0.37 \\ \cmidrule{3-6}
FC & $D_4$ & 96.25$\pm$0.21 & 5.36$\pm$0.73 & 96.30$\pm$0.30 & 5.88$\pm$1.31 \\ \cmidrule{3-6}
S-FC &  & 96.87$\pm$0.66 & 21.60$\pm$1.01 & 96.39$\pm$0.71 & 21.35$\pm$1.14 \\ \cmidrule{3-6}
MTAE &  & 1.23 & 5 & 1.3275 & 5.18 \\ \cmidrule{3-6}
S-MTAE &  & \textbf{99.95} & \textbf{23.21} & \textbf{99.88} & \textbf{22.00} \\ \cmidrule{3-6}
 &  &  &  &  &  \\ \cmidrule{3-6}
AGG & & 61.37$\pm$0.56 & 5.36$\pm$0.53 & 61.70$\pm$0.44 & 5.73$\pm$0.64 \\ \cmidrule{3-6}
S-AGG &  & 66.90$\pm$0.81 & \textbf{20.06$\pm$0.24} & 67.37$\pm$0.83 & \textbf{19.75$\pm$0.51} \\ \cmidrule{3-6}
FC & $D_5$ & 64.74$\pm$1.72 & 5.48$\pm$0.80 & 64.53$\pm$0.86 & 5.81$\pm$1.55 \\ \cmidrule{3-6}
S-FC &  & 63.88$\pm$2.82 & 18.48$\pm$1.34 & 64.46$\pm$2.39 & 19.26$\pm$0.60 \\ \cmidrule{3-6}
MTAE &  & 1.567 & 5 & 1.25 & 5.48 \\ \cmidrule{3-6}
S-MTAE &  & \textbf{78.52} & 19.21 & \textbf{78.81} & 19.23 \\
\hline
\end{tabular}
}
\caption{Domain Generalization (DG) and Zero-Shot Domain Generalization (ZSDG) performance of different domains on \textbf{CIFAR-100} dataset.}
\label{tab:full_cifar_100}
\end{table}

\begin{table}[]
\resizebox{\textwidth}{!}{%
\begin{tabular}{lccccccccc}
\hline
& TARGET &\multicolumn{2}{c}{Setting 1}& \multicolumn{2}{c}{Setting 2}& \multicolumn{2}{c}{Setting 3}& \multicolumn{2}{c}{Setting 4}\\
\cmidrule{3-10}
&  & DG & ZSDG & DG & ZSDG & DG & ZSDG& DG & ZSDG \\
\cmidrule{3-10}

AGG & & 91.23$\pm$0.27 & 43.23$\pm$13.95 & 88.30$\pm$0.85 & 60.17$\pm$3.20 & 87.79$\pm$0.45 & 56.20$\pm$24.44 & 87.06$\pm$0.33 & 50.37$\pm$14.73 \\
\cmidrule{3-10}
S-AGG & P & 88.61$\pm$0.93 & \textbf{87.81$\pm$2.82} & 83.11$\pm$0.84 & \textbf{88.92$\pm$1.50} & 82.35$\pm$0.73 & \textbf{67.51$\pm$1.40} & 82.79$\pm$0.77 & \textbf{71.46$\pm$1.53} \\
\cmidrule{3-10}
FC &  & \textbf{91.60$\pm$0.57} & 31.04$\pm$6.31 & \textbf{89.10$\pm$0.52} & 52.35$\pm$10.79 & \textbf{88.51$\pm$0.66} & 59.12$\pm$24.60 & \textbf{87.48$\pm$0.19} & 46.32$\pm$12.79 \\
\cmidrule{3-10}
S-FC &  & 87.48$\pm$1.14 & 81.95$\pm$1.59 & 82.25$\pm$0.87 & 82.35$\pm$3.54 & 81.95$\pm$1.46 & 58.07$\pm$11.15 & 82.37$\pm$1.08 & 67.30$\pm$1.67 \\
\cmidrule{3-10}
 &  &  &  &  &  &  &  &  &  \\
\cmidrule{3-10}
AGG & & \textbf{65.03$\pm$1.10} & 57.81$\pm$12.33 & 65.85$\pm$0.69 & 48.06$\pm$3.65 & 73.53$\pm$0.63 & 54.07$\pm$7.33 & 64.14$\pm$0.39 & 53.73$\pm$7.86 \\
\cmidrule{3-10}
S-AGG & A & 58.82$\pm$1.15 & \textbf{58.40$\pm$4.87} & 60.67$\pm$0.88 & \textbf{61.90$\pm$0.76} & 69.91$\pm$1.47 & 36.94$\pm$2.01 & 59.73$\pm$1.08 & 49.29$\pm$0.96 \\
\cmidrule{3-10}
FC &  & 64.73$\pm$0.84 & 48.84$\pm$17.75 & \textbf{66.42$\pm$0.63} & 46.93$\pm$5.99 & \textbf{75.21$\pm$0.89} & \textbf{57.79$\pm$10.56} & \textbf{65.21$\pm$1.29} & \textbf{53.96$\pm$6.93} \\
\cmidrule{3-10}
S-FC &  & 52.18$\pm$1.07 & 55.68$\pm$2.93 & 55.97$\pm$1.21 & 61.78$\pm$2.05 & 63.25$\pm$2.31 & 52.55$\pm$2.39 & 54.38$\pm$0.88 & 47.20$\pm$1.31 \\
\cmidrule{3-10}
 &  &  &  &  &  &  &  &  &  \\
\cmidrule{3-10}
AGG & & \textbf{70.89$\pm$0.94} & 50.39$\pm$9.42 & \textbf{79.13$\pm$1.10} & 57.86$\pm$13.64 & \textbf{72.40$\pm$1.15} & \textbf{60.05$\pm$6.54} & 69.13$\pm$1.07 & 44.80$\pm$15.57 \\
\cmidrule{3-10}
S-AGG & C & 68.48$\pm$1.36 & \textbf{82.30$\pm$1.70} & 75.55$\pm$1.05 & \textbf{70.48$\pm$3.16} & 71.62$\pm$1.71 & 50.93$\pm$1.76 & 68.61$\pm$0.76 & \textbf{60.11$\pm$4.13} \\
\cmidrule{3-10}
FC &  & 67.50$\pm$1.25 & 48.62$\pm$11.79 & 78.79$\pm$0.74 & 53.95$\pm$7.99 & 71.66$\pm$0.89 & 57.91$\pm$5.27 & \textbf{69.16$\pm$0.22} & 41.07$\pm$16.56 \\
\cmidrule{3-10}
S-FC &  & 64.77$\pm$1.26 & 74.35$\pm$4.52 & 69.93$\pm$2.19 & 65.67$\pm$3.03 & 66.61$\pm$1.51 & 54.02$\pm$2.81 & 63.47$\pm$1.57 & 53.35$\pm$5.87 \\
\cmidrule{3-10}
 &  &  &  &  &  &  &  &  &  \\
\cmidrule{3-10}
AGG & & \textbf{70.17$\pm$1.22} & 40.72$\pm$42.45 & 76.31$\pm$1.21 & 37.52$\pm$25.91 & \textbf{68.21$\pm$1.41} & 51.57$\pm$28.41 & \textbf{66.24$\pm$2.00} & 27.36$\pm$24.88 \\
\cmidrule{3-10}
S-AGG & S & 63.93$\pm$1.27 & 92.02$\pm$0.66 & 69.17$\pm$1.44 & \textbf{90.47$\pm$1.55} & 61.10$\pm$2.67 & 45.47$\pm$9.09 & 61.69$\pm$0.91 & \textbf{86.52$\pm$2.86} \\
\cmidrule{3-10}
FC &  & 67.89$\pm$2.91 & 53.63$\pm$33.71 & \textbf{76.76$\pm$2.04} & 32.36$\pm$34.76 & 67.37$\pm$1.24 & 54.66$\pm$28.20 & 65.04$\pm$0.87 & 33.04$\pm$31.48 \\
\cmidrule{3-10}
S-FC &  & 60.02$\pm$1.79 & \textbf{92.92$\pm$1.69} & 62.06$\pm$2.16 & 90.11$\pm$4.09 & 58.62$\pm$1.99 & \textbf{69.17$\pm$9.07} & 55.90$\pm$2.69 & 84.49$\pm$6.75 \\
\hline
\end{tabular}}
\caption{Domain Generalization (DG) and Zero-Shot Domain Generalization (ZSDG) performance on \textbf{PACS}
% \textbf{PACS(Picture, Art, Cartoon, Sketch)} 
Dataset.}
\label{tab:full_pacs}
\end{table}

\end{document}